%% file: acl2020.tex
\documentclass[11pt,a4paper]{article}
\usepackage[hyperref]{acl2020}
\usepackage{times}
\usepackage{latexsym}

\usepackage{microtype}

\usepackage{graphicx}
\usepackage{url}
\usepackage{bm}
\usepackage{amssymb,amsfonts,amsmath}
\usepackage{algorithm}
\usepackage{algorithmicx}
\usepackage{algpseudocode}
\usepackage{mathtools}
\usepackage{booktabs}
\usepackage{subcaption}
\usepackage{xspace}
\usepackage{dcolumn}
\usepackage{comment}
\usepackage{multirow}
\usepackage{enumitem}

\aclfinalcopy 

\newcommand\xlmt{XLM-T}

\title{\xlmt{}: Scaling up Multilingual Machine Translation with \\ Pretrained Cross-lingual Transformer Encoders}

\author{
 Shuming Ma, 
 Jian Yang,
 Haoyang Huang,
 Zewen Chi,
 Li Dong, \\
 \textbf{Dongdong Zhang,}
 \textbf{Hany Hassan Awadalla,}
 \textbf{Alexandre Muzio,}
 \textbf{Akiko Eriguchi,} \\
 \textbf{Saksham Singhal,}
 \textbf{Xia Song,}
 \textbf{Arul Menezes,}
 \textbf{Furu Wei}
 \\
 Microsoft Corporation\\
 \texttt{\{shumma,t-jianya,haohua,v-zechi,lidong1,dozhang\}@microsoft.com} \\
 \texttt{\{hanyh,alferre,akikoe,saksingh,xiaso,arulm,fuwei\}@microsoft.com} }

\date{}

\begin{document}
\maketitle
\begin{abstract}
Multilingual machine translation enables a single model to translate between different languages. Most existing multilingual machine translation systems adopt a randomly initialized Transformer backbone. In this work, inspired by the recent success of language model pre-training, we present \xlmt{}, which initializes the model with an \textit{off-the-shelf} pretrained cross-lingual Transformer encoder and fine-tunes it with multilingual parallel data. This simple method achieves significant improvements on a WMT dataset with 10 language pairs and the OPUS-100 corpus with 94 pairs. Surprisingly, the method is also effective even upon the strong baseline with back-translation. Moreover, extensive analysis of \xlmt{} on unsupervised syntactic parsing, word alignment, and multilingual classification explains its effectiveness for machine translation.\footnote{The code will be at \url{https://aka.ms/xlm-t}.}
\end{abstract}

\section{Introduction}

Multilingual neural machine translation (NMT) enables a single model to translate between multiple language pairs, which has drawn increasing attention in the community~\cite{multiway,haea2016,googlemnmt,mmnmt,m2m}. Recent work shows that multilingual machine translation achieves promising results especially for low-resource and zero-resource machine translation~\cite{zeroresource,lowresource,unsupervised,opus100}.

\begin{figure}[t]
\centering
\includegraphics[width=1.0\linewidth]{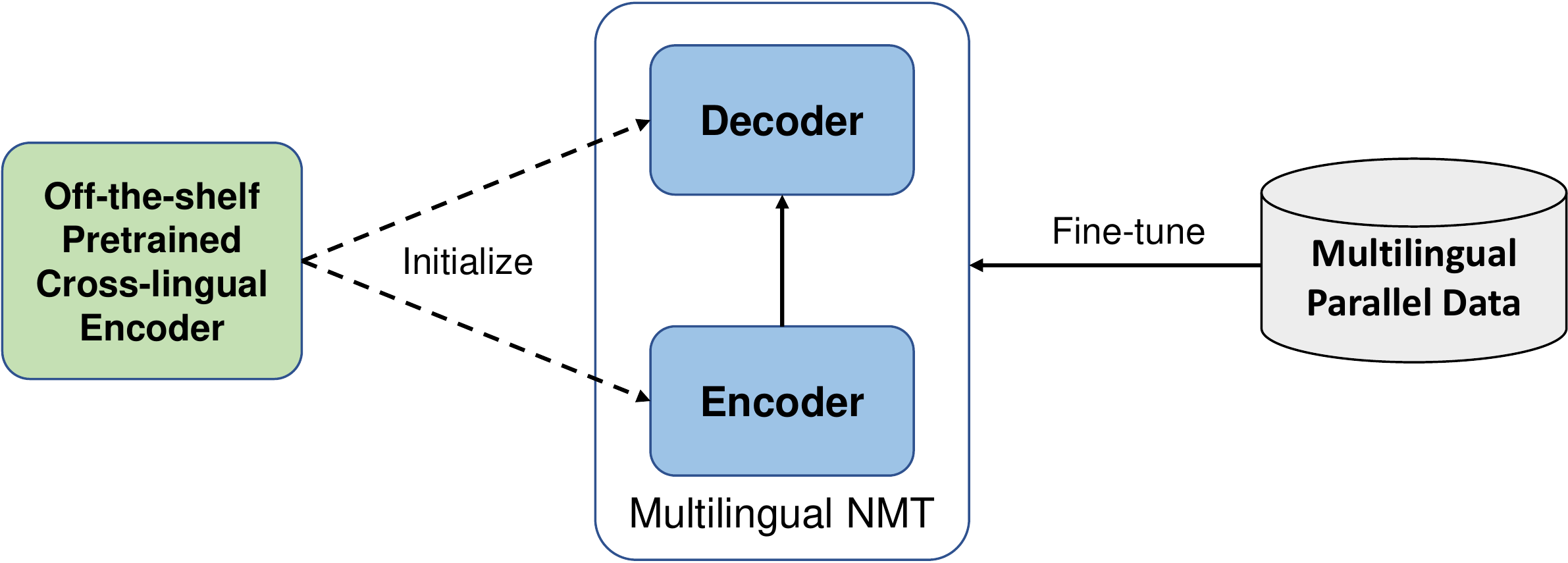}
\caption{Framework of \xlmt{}. We use off-the-shelf pretrained cross-lingual encoders (such as XLM-R) to initialize both the encoder and decoder of the multilingual NMT model. Then we fine-tune the model on multilingual parallel data.}
\end{figure}

Pre-training-then-fine-tuning framework~\cite{bert,roberta,unilm,mass,t5} has shown substantial improvements on many natural language processing (NLP) tasks by pre-training a model on a large corpus and fine-tuning it on the downstream tasks. Pre-training multilingual language models~\cite{xlm,xlmr,xnlg,infoxlm,mt5} obtains significant performance gains on a wide range of cross-lingual tasks, which is naturally applicable to multilingual machine translation where the representations are shared among different languages. Moreover, pre-training has great potential in efficiently scaling up multilingual NMT, while existing methods, such as back-translation~\cite{bt}, are expensive in the multilingual setting.

Most existing work~\cite{xlm,mass,bart} on leveraging pretrained models for machine translation mainly lies in the bilingual setting. How to effectively and efficiently use these existing pretrained models for multilingual machine translation is not fully explored. \citet{mbart} introduce a sequence-to-sequence denoising auto-encoder (mBART)  pretrained on large-scale monolingual corpora in many languages.
\citet{mrasp} propose to pretrain the multilingual machine translation models with a code-switching objective function. However, this model requires a large-scale parallel data for pre-training, which hinders its application to low-resource and zero-resource languages.

In this work, we present a simple and effective method \xlmt{} that initializes multilingual machine translation with a pretrained cross-lingual Transformer encoder and fine-tunes it using multilingual parallel data.
The cross-lingual pretrained encoders are \textit{off-the-shelf} for general cross-lingual NLP tasks so we do not need to specifically pretrain for machine translation. We adopt XLM-R~\cite{xlmr} as the pretrained encoder and conduct extensive experiments on multilingual machine translation with 10 language pairs from WMT datasets\footnote{http://www.statmt.org} and 94 language pairs from OPUS datasets\footnote{http://opus.nlpl.eu/opus-100.php}. This simple method achieves significant and consistent gains on both large-scale datasets. The improvement is still significant over the strong baseline with back-translation. 

To analyze how the pretrained encoders benefit multilingual machine translation, we perform some probing tasks for both \xlmt{} and a randomly initialized multilingual NMT baseline. Empirical studies show that \xlmt{} improves the abilities of syntactic parsing, word alignment, and multilingual classification. We believe that this work can shed light on further improvements of applying pretrained models to machine translation.

\section{\xlmt{}}

In this section, we introduce our proposed model: Cross-lingual Language Modeling Pre-training for Translation, which is denoted as \xlmt{}.

\subsection{Multilingual Machine Translation}

Suppose we have $L$ languages to translate in a model. Among these languages, we have $N$ bilingual corpora, each of which contains parallel sentences $\{(x_{L_i}^1, x_{L_j}^1), \cdots, (x_{L_i}^k, x_{L_j}^k)\}$ between $L_i$ and $L_j$, where $k$ is the number of training instances.

Given the corpora, we are able to train a multilingual model $\mathbf{P}_{\theta}$ that enables the translation among different languages. With the parallel data of $N$ language direction, the model is learnt with a combination of different objective:

\begin{equation}
    L=-\sum_{i,j,k} \log \mathbf{P}_{\theta}(x_{L_i}^k, x_{L_j}^k)\label{objective}
\end{equation}

Typically, the multilingual NMT model uses a unified model that shares the encoders and decoders for all translation directions. In this work, we adopt the state-of-the-art Transformer as the backbone model $\mathbf{P}_{\theta}$. Following the methods of~\citet{haea2016} and~\citet{googlemnmt}, we prepend a target language token to each source sentence to indicate which language should be translated on the target side.

\subsection{Cross-lingual Pretrained Encoders}

In this work, we argue that multilingual NMT models can be scaled up by pre-training the encoder with large-scale monolingual data. Multilingual NMT encourages a shared representation among different languages so that the data in one language helps to model the other language. Meanwhile, cross-lingual pretrained encoders prove to be effective in transferring cross-lingual representations. 

In this work, we adopt XLM-R \textsc{base}~\cite{xlmr} as the pretrained encoder. It was trained in 100 languages, using more than two terabytes of filtered CommonCrawl data. XLM-R is based on the Transformer architecture, trained using the multilingual masked language model (MLM) objective~\cite{xlm}. It has a shared vocabulary of 250,000 tokens based on SentencePiece model~\cite{sentencepiece}.

\subsection{Initialization Strategy}

Given the above pretrained encoder, we can use it to initialize the encoder and decoder of the Transformer-based multilingual NMT model. 

\paragraph{Initializing cross-lingual encoder}

There are different Transformer variants in terms of the NMT encoder. To initialize our NMT encoder with pretrained XLM-R, we make their architectures consistent.
We add a layer normalization layer after the embedding layer and do not scale up the word embedding.
We use post layer normalization for both the attention layers and feed-forward layers. The activation insides the feed-forward layers is GELU~\cite{gelu}. The positional embedding is learned during training.

\paragraph{Initializing cross-lingual decoder}

The pretrained encoder can be also used to initialize the decoder. The architecture of the decoder is the same as that of the encoder, except that there is a cross-attention layer after the self-attention layer. Due to this difference, we explore several methods to initialize the decoder, including sharing the weights of cross-attention layers and self-attention layers and randomly initializing the cross-attention.

\subsection{Multilingual Fine-tuning}

We can now fine-tune our \xlmt{} model with the objective function (Eq.~\ref{objective}). 
A simple concatenation of all parallel data will lead to poor performance on low-resource translation because of the imbalanced data.
Following the previous work~\cite{mmnmt,zcode}, we adopt a temperature-based batch balance method by sampling the sentence pairs in different languages according to a multinomial distribution with probabilities $\{q_1, q_2, \cdots, q_N\}$:
\begin{equation}
    q_{i}=\frac{p_{i}^{\frac{1}{T}}}{\sum_{j=1}^{L} p_{j}^{\frac{1}{T}}}
\end{equation}
\begin{equation}
    p_{i}=\frac{|L_{i}|}{\sum_{L} |L_{j}|}
\end{equation}
where $N$ is the number of translation directions, $|L_{i}|$ is the number of parallel data for $i$-th direction, and $T$ is a temperature.

To reduce over-sampling of low-resource languages in the early stage of training, we employ a dynamic temperate sampling mechanism~\cite{zcode}. The temperature is low at the beginning of training and is gradually increased for the first several epochs. Formally, the temperature can be written as:
\begin{equation}
    T_{i}=\min(T, T_{0}+\frac{i}{N}(T-T_{0}))
\end{equation}
where $T_0$ is the initial temperature, $T$ is the peak temperature, and $N$ is the number of warming-up epochs.
For a fair comparison, we set $T_0=1.0$, $T=5.0$, and $N=5$ for all the experiments in our work.

\section{Experimental Setup}

\subsection{Data}

\paragraph{WMT-10} 

Following~\cite{zcode}, we use a collection of parallel data in different languages from the WMT datasets to evaluate the models. The parallel data is between English and other 10 languages, including French (Fr), Czech (Cs), German (De), Finnish (Fi), Latvian (Lv), Estonian (Et), Romanian (Ro), Hindi (Hi), Turkish (Tr) and Gujarati (Gu). We choose the data from the latest available year of each language and exclude WikiTiles. We also remove the duplicated samples and limit the number of parallel data in each language pair up to 10 million by randomly sampling from the whole corpus. We use the same test sets and validation set as in~\cite{zcode}. The details can be found in Appendix.

In the back-translation setting, we collect large-scale monolingual data for each language from NewsCrawl\footnote{http://data.statmt.org/news-crawl}. We remove the data with low quality, and randomly sample 5 million sentences in each language. For the languages without enough data (Fi, Lv, Et, Gu), we also sample additional data from CCNet~\cite{ccnet} to combine with that from NewsCrawl. We use a target-to-source multilingual NMT model to back-translate these monolingual data as the augmented parallel data.

\paragraph{OPUS-100}

To evaluate our model in the massively multilingual machine translation setting, we use the OPUS-100 corpus provided by~\citet{opus100}. 
OPUS-100 is an English-centric multilingual corpus covering 100 languages, which is randomly sampled from the OPUS collection.

The dataset is split into training, development, and test sets. The training set has up to 1 million sentence pairs per language pair, while the development and test sets contain up to 2000 parallel sentences.
The whole dataset contains approximately 55 million sentence pairs. We remove 5 languages without any development set or test sets, which results in 95 languages including English.

\begin{table*}[t]
\centering
\begin{tabular}{l|cccccccccc|c}
\toprule
X $\rightarrow$ En & Fr & Cs & De & Fi & Lv & Et & Ro & Hi & Tr & Gu & Avg \\
\midrule
\multicolumn{12}{l}{\textit{Train on Original Parallel Data (Bitext)}} \\
\midrule
Bilingual NMT & 36.2 & 28.5 & 40.2 & 19.2 & 17.5 & 19.7 & 29.8 & 14.1 & 15.1 & 9.3 & 23.0 \\
\midrule
Many-to-One & 34.8 & 29.0 & 40.1 & 21.2 & 20.4 & 26.2 & 34.8 & 22.8 & 23.8 & 19.2 & 27.2  \\
\xlmt{} & 35.9 &  30.5 &  41.6 &  22.5 & 21.4 &  28.4 &  36.6 & 24.6 &  25.6 & 20.4 & \bf 28.8 \\
\midrule
Many-to-Many & 35.9 & 29.2 & 40.0 & 21.1 & 20.4 & 26.3 & 35.5 & 23.6 & 24.3 &  20.6 & 27.7 \\
\xlmt{} & 35.5 & 30.0 & 40.8 & 22.1 &  21.5 & 27.8 & 36.5 &  25.3 & 25.0 &  20.6 & \bf 28.5 \\
\midrule
\multicolumn{12}{l}{\textit{Train on Original Parallel Data and \textbf{Back-Translation} Data (Bitext+BT)}} \\
\midrule
\cite{zcode} & 35.3 & 31.9 & 45.4 & 23.8 & 22.4 & 30.5 & 39.1 & 28.7 & 27.6 & 23.5 & 30.8 \\
Many-to-One  & 35.9 & 32.6 & 44.1 & 24.9 & 23.1 & 31.5 & 39.7 & 28.2 & 27.8 & 23.1 & 31.1 \\
\xlmt{}  &  36.0 &  33.1 &  44.8 &  25.4 &  23.9 &  32.7 &  39.8 &   30.1 &  28.8 &  23.6 & \bf 31.8 \\
\midrule
\cite{zcode} & 35.3 & 31.2 & 43.7 & 23.1 & 21.5 & 29.5 & 38.1 & 27.5 & 26.2 & 23.4 & 30.0 \\
Many-to-Many & 35.7 & 31.9 & 43.7 & 24.2 & 23.2 & 30.4 & 39.1 & 28.3 & 27.4 & 23.8 & 30.8 \\
\xlmt{} &  36.1 & 32.6 & 44.3 &  25.4 & 23.8 & 32.0 &  40.3 & 29.5 & 28.7 &  24.2 & \bf 31.7 \\
\bottomrule
\end{tabular}
\caption{X $\rightarrow$ En test BLEU for bilingual, many-to-one, and many-to-many models on WMT-10. On the top are the models trained with original parallel data, while the bottom are combined with back-translation. The languages are ordered from high-resource (left) to low-resource (right).}
\label{table:x2e}
\end{table*}

\begin{table*}[t]
\centering
\begin{tabular}{l|cccccccccc|c}
\toprule
En $\rightarrow$ X & Fr & Cs & De & Fi & Lv & Et & Ro & Hi & Tr & Gu & Avg \\
\midrule
\multicolumn{12}{l}{\textit{Train on Original Parallel Data (Bitext)}} \\
\midrule
Bilingual NMT & 36.3 & 22.3 & 40.2 & 15.2 & 16.5 & 15.0 & 23.0 & 12.2 & 13.3 & 7.9 & 20.2\\
\midrule
One-to-Many & 34.2 & 20.9 & 40.0 & 15.0 & 18.1 & 20.9 & 26.0 & 14.5 & 17.3 & 13.2 & 22.0 \\
\xlmt{} & 34.8 & 21.4 & 39.9 & 15.4 & 18.7 & 20.9 & 26.6 & 15.8 & 17.4 & 15.0 & \bf 22.6 \\
\midrule
Many-to-Many & 34.2 & 21.0 & 39.4 & 15.2 & 18.6 & 20.4 & 26.1 & 15.1 & 17.2 & 13.1 & 22.0 \\
\xlmt{} & 34.2 & 21.4 & 39.7 & 15.3 & 18.9 & 20.6 & 26.5 & 15.6 & 17.5 & 14.5 & \bf 22.4 \\
\midrule
\multicolumn{12}{l}{\textit{Train on Original Parallel Data and \textbf{Back-Translation} Data (Bitext+BT)}} \\
\midrule
\cite{zcode} & 36.1 & 23.6 & 42.0 & 17.7 & 22.4 & 24.0 & 29.8 & 19.8 & 19.4 & 17.8 & 25.3 \\
One-to-Many & 36.8 & 23.6 & 42.9 &  18.3 & 23.3 & 24.2 & 29.5 &  20.2 & 19.4 &  13.2 & 25.1 \\
\xlmt{} &  37.3 &  24.2 &  43.6 & 18.1 &  23.7 &  24.2 &  29.7 & 20.1 &  20.2 & 13.7 & \bf  25.5 \\
\midrule
\cite{zcode} & 35.8 & 22.4 & 41.2 & 16.9 & 21.7 & 23.2 & 29.7 & 19.2 & 18.7 & 16.0 & 24.5 \\
Many-to-Many & 35.9 & 22.9 & 42.2 & 17.5 & 22.5 & 23.4 & 28.9 & 19.8 & 19.1 & 14.5 & 24.7 \\
\xlmt{} & 36.6 & 23.9 & 42.4 & 18.4 & 22.9 & 24.2 & 29.3 & 20.1 & 19.8 & 12.8 & \bf 25.0 \\
\bottomrule
\end{tabular}
\caption{En $\rightarrow$ X test BLEU for bilingual, many-to-one, and many-to-many models on WMT-10. On the top are the models trained with original parallel data, while the bottom are combined with back-translation. The languages are ordered from high-resource (left) to low-resource (right).}
\label{table:e2x}
\end{table*}

\begin{table*}[t]
\centering
\small
\begin{tabular}{l|ccccc|ccccc}
\toprule
\multirow{2}{*}{Models} & \multicolumn{5}{c|}{X $\rightarrow$ En} & \multicolumn{5}{c}{En $\rightarrow$ X}  \\
\cmidrule(l){2-11}
& High & Med & Low & Avg & WR  & High & Med & Low & Avg & WR   \\
\midrule
Best System from~\cite{opus100} & 30.3 & 32.6 & 31.9 & 31.4 & - & 23.7 & 25.6 & 22.2 & 24.0 & - \\
\midrule
Many-to-Many & 31.5 & 35.1 & 36.0 & 33.6 & \textit{ref} & 25.6 & 30.5 & 30.5 & 28.2& \textit{ref}  \\
\xlmt{} & \bf 32.4 & \bf 35.9 & \bf 36.9 & \bf 34.5 & \bf 89.4 & \bf 26.1 & \bf 30.9 & \bf 31.0 & \bf 28.6 & \bf 75.5 \\
\bottomrule
\end{tabular}
\caption{X $\rightarrow$ En and En $\rightarrow$ X test BLEU for high/medium/low resource language pairs in many-to-many setting on OPUS-100 test sets. The BLEU scores are average across all language pairs in the respective groups. ``WR'': win ratio (\%) compared to \textit{ref}.}
\label{table:opus}
\end{table*}

\subsection{Pretrained Models and Baselines}

We use the state-of-the-art Transformer model for all our experiments with the fairseq\footnote{https://github.com/pytorch/fairseq} implementation~\cite{fairseq}. For the baseline model of the WMT-10 dataset, we adopt a Transformer-big architecture with a 6-layer encoder and decoder. The hidden size, embedding size and the number of attention head is 1024, 1024, and 16 respectively, while the dimension of feedforward layer is 4096. We tokenize the data with SentencePiece model~\cite{sentencepiece} with a vocabulary size of 64,000 tokens extracted from the training set. 

For \xlmt{}, we initialize with XLM-R base model, which has 12-layer encoder, 6-layer decoder, 768 hidden size, 12 attention head, and 3,072 dimensions of feedforward layers. We do not use a deeper decoder because our preliminary experiments show no improvement by increasing the number of decoder layers, which is consistent with the observations in ~\cite{deepencoder}. Different from WMT-10, massively multilingual NMT suffers from weak capacity~\cite{opus100}. Therefore, for the baseline of the OPUS-100 dataset, we adopt the same architecture and vocabulary as \xlmt{} but randomly initializing the parameters so that the numbers of parameters are the same. We tie the weights of encoder embeddings, decoder embeddings, and output layers in all experiments.

\subsection{Training and Evaluation}

We train all models with Adam Optimizer~\cite{adam} with $\beta_{1}=0.9$ and $\beta_{2}=0.98$. The learning rate is among \{3e-4, 5e-4\} with a warming-up step of 4,000. The models are trained with the label smoothing cross-entropy, and the smoothing ratio is 0.1. We set the dropout of attention layers as 0.0, while the rest of the dropout rate is 0.1. We limit the source length and the target length to be 256. For the WMT-10 dataset, the batch size is 4,096 and we accumulate the gradients by 16 batches. For the OPUS-100 dataset, we set the batch size as 2,048 and the gradients are updated every 32 batches. All experiments on the WMT-10 dataset are conducted on 8 V100 GPUs, while the experiments on OPUS-100 are on a DGX-2 machine with 16 V100 GPUs. 

During testing, we use the beam search algorithm with a beam size of 5. We set the length penalty as 1.0. The last 5 checkpoints are averaged for evaluation. We report the case-sensitive detokenized BLEU using sacreBLEU\footnote{BLEU+case.mixed+lang.\{src\}-\{tgt\}+numrefs.1+smooth.exp+tok.13a+version.1.4.14}~\cite{sacrebleu}.

\section{Results}

\subsection{WMT-10}

We study the performance of \xlmt{} in three multilingual translation scenarios, including many-to-English (X $\rightarrow$ En), English-to-many (En $\rightarrow$ X), and many-to-many (X $\rightarrow$ Y). For many-to-many, we use a combination of English-to-many and many-to-English as the training data. We compare \xlmt{} with both the bilingual NMT and the multilingual NMT models to verify the effectiveness.

Table~\ref{table:x2e} reports the results on the X $\rightarrow$ En test sets. Compared with the bilingual baseline, the multilingual models achieve much better performance on the low-resource languages and are worse on the high-resource languages. In general, the multilingual baseline outperforms the bilingual baselines by an average of +4.2 points.
In the many-to-English scenario, \xlmt{} achieves significant improvements over the multilingual baseline across all 10 languages. The average gain is +1.6 points.
In the many-to-many scenario, the gain becomes narrow, but still reaches +0.8 points over the multilingual baseline.
We further combine the parallel data with back-translation. Back-translation results in a large gain of +3.9 BLEU score over the baseline. Therefore, back-translation is a strong baseline for multilingual NMT. In the back-translation setting, \xlmt{} can further improve this strong baseline by a significant gain of +0.7 points, showing the effectiveness of \xlmt{}. As for the many-to-many setting, the improvement is even larger, reaching a difference of +0.9 points. We compare \xlmt{} with \citet{zcode}'s method. Besides back-translation, they use the monolingual data (i.e. the target side of back-translation data) with two tasks of Mask Language Model (MLM) and Denoising AutoEncoder (DAE). It shows that \xlmt{} can outperform this method in both the many-to-one and many-to-many settings.

Table~\ref{table:e2x} summarizes the results on the En $\rightarrow$ X test sets. Similar to the results of X $\rightarrow$ En, the multilingual NMT improves the average BLEU score of the bilingual baseline, while \xlmt{} beats the multilingual baseline by +0.6 points. As for the many-to-many and back-translation scenarios, \xlmt{} yields the increments of +0.4 points, +0.4 points, and +0.3 points, respectively. Compared with \citet{zcode}'s method, \xlmt{} has similar performance in the one-to-many setting, and a slightly improvement of +0.5 BLEU in the many-to-many scenario. The performance of \xlmt{} in Gu is worse than that of \citet{zcode}. We conjecture that this is related to the implementation details of data sampling. Generally, the improvements are smaller than X $\rightarrow$ En. We believe it is because the multilingual part of En $\rightarrow$ X is at the decoder side, which XLM-R is not an expert in. How to improve En $\rightarrow$ X with pretrained models is a promising direction to explore in the future.

\subsection{OPUS-100}

To further verify the effectiveness of \xlmt{} on massively multilingual machine translation, we conduct experiments on OPUS-100, which consists of 100 languages including English. After removing 5 languages without test sets, we have 94 language pairs from and to English. Following~\citet{opus100}, we group the languages into three categories, including high-resource languages ($\geq$0.9M, 45 languages), low-resource languages ($<$0.1M, 21 languages), and medium-resource languages (the rest, 28 languages). According to the previous work~\cite{opus100}, the performance of massively multilingual machine translation is sensitive to the model size (i.e. the number of parameters), because the model capacity is usually the bottleneck when the numbers of languages and data are massive. Therefore, we make the architectures of baseline and \xlmt{} consistent to ensure the parameters are exactly equal.

Both the multilingual baseline and \xlmt{} are trained in the many-to-many setting.
Table~\ref{table:opus} reports their results on OPUS-100 as well as the performance of the best system from~\citet{opus100}. For the X $\rightarrow$ En test sets, \xlmt{} has consistent and significant gains over the multilingual baseline for all the high (+0.9 BLEU), medium (+0.8 BLEU), and low (+0.9 BLEU) resource languages. The overall improvement is +0.9 points by averaging all 94 En $\rightarrow$ X language pairs. For the En $\rightarrow$ X test sets, \xlmt{} also benefits the high/medium/low resource languages. Generally, the performance improves by +0.4 points in terms of the average BLEU scores on En $\rightarrow$ X test sets. We also compute the win ratio (WR), which counts the proportion of languages where \xlmt{} outperforms the baselines. It shows that \xlmt{} is better in 89.4\% of the language pairs on the X $\rightarrow$ En test sets and 75.5\% on the En $\rightarrow$ X test sets.

\begin{table}[t]
\centering
\small
\begin{tabular}{l|cc|c}
\toprule
Models & \#Layer & \#Hidden & BLEU \\
\midrule
Multilingual NMT & 6/6 & 1024 & 27.2 \\
Multilingual NMT & 12/6 & 768 & 26.9 \\
\xlmt{} & 12/6 & 768 & \bf 28.8 \\
\bottomrule
\end{tabular}
\caption{Ablation study of Transformer architectures on WMT-10 test sets. The BLEU scores are average across 10 languages on WMT-10 X $\rightarrow$ En test sets. \#Layer denotes the number of encoder/decoder layers, while \#Hidden means the hidden size.}
\label{table:architecture}
\end{table}

\begin{table}[t]
\centering
\small
\begin{tabular}{l|ccc}
\toprule
Models & X $\rightarrow$ En & En $\rightarrow$ X \\
\midrule
Multilingual NMT & 27.7 & 22.0 \\
\xlmt{} (enc.) & 28.4 & 22.0  \\
\xlmt{} (enc.+dec.) & \bf 28.5 & \bf 22.4 \\
\bottomrule
\end{tabular}
\caption{Ablation study on different initialization strategies in the many-to-many setting on WMT-10 test sets. The BLEU scores are average on each test set.}
\label{table:init}
\end{table}

\subsection{Ablation Studies}

\paragraph{Effect of architectures}

For the WMT-10 experiments, the architecture of \xlmt{} is different from the multilingual baseline, including the number of encoder layers, the hidden size, the layer normalization layer, and the activation function. To identify whether the architecture or the weights of \xlmt{} improves the performance, we perform an ablation study by initializing \xlmt{} with random weights.
Table~\ref{table:architecture} shows that the architecture of \xlmt{} does not improve the performance of the multilingual baseline, leading to a slight drop of -0.3 points. With our initialization strategies, \xlmt{} improves by a significant gain of +1.9 points. This proves that the initialization of \xlmt{} is the main contribution of the improvement. For the OPUS-100 experiments, the architecture of \xlmt{} is the same as the multilingual baseline, so we do not need any additional ablation on the architecture.

\paragraph{Effect of initialization strategies}

To analyze the effect of the proposed initialization strategies, we conduct an ablation study by removing the encoder initialization and decoder initialization. Table~\ref{table:init} summarizes the results. It shows that the encoder initialization mainly contributes to the improvements of X $\rightarrow$ En. It is because that the source sides of this scenario are multilingual, while that of E $\rightarrow$ X is English-only. Similarly, the decoder initialization mainly benefits E $\rightarrow$ X, whose target side is multilingual. Moreover, it concludes that the encoder initialization contributes to more gains than the decoder initialization for multilingual NMT. The reason may be XLM-R is more consistent with the encoder, while lacks the modeling of cross-attention layer for the decoder.

\section{Analysis}

To analyze how \xlmt{} improves multilingual machine translation, we perform three probing tasks, including unsupervised dependency parsing, multilingual classification, and word alignment retrieval.

\subsection{Word Alignment}

Word alignment is an important metric to evaluate the ability to transfer between different languages. We assume that \xlmt{} improves the internal translation transfer by improving the similarity of encoder representations between two translated words. Therefore, the ability to translate one language can easily benefit that of translating the other language. To evaluate the performance of word alignment, we use the same labeled alignment data as in~\cite{simalign}, which is original from Europarl and WPT datasets. The alignment data is between English and six other languages, including Czech, German, French, Hindi, Romanian, and Persian. We discard Persian and Hindi, which is either not in WMT-10 or only contains 90 test samples.

\paragraph{Setup}

We compare the alignment error rate between \xlmt{} and multilingual NMT baseline. Both models are trained with the WMT-10 dataset in the many-to-many scenario. Given a sentence pair, we prepend a language token to each sentence and compute the representations of each word by averaging the representations of its subwords. A similarity matrix can be obtained by calculating the cosine distance between words from two sentences. With the similarity matrices, we use the IterMax~\cite{simalign} algorithm to extract the alignments. IterMax is iterative Argmax, which modifies the similarity matrix conditioned on the alignment edges found in a previous iteration. We compare the extracted alignments with the ground truth to measure the alignment error rate.

\begin{table}[t] 
\centering
\small
\scalebox{0.88}{
\begin{tabular}{l|cccc|c}
\toprule
Models & Cs & De & Fr & Ro & Avg \\
\midrule
XLM-R &30.78 & 26.46&26.24 &31.74 &28.81 \\
\midrule
Multilingual NMT &24.16  &21.37  &31.18  &28.90 &26.40\\
\xlmt{}        &20.97  &21.47  &30.89 &24.91 &\bf 24.56 \\
\bottomrule
\end{tabular}}
\caption{Analysis of word alignment error on \citet{simalign}'s alignment datasets. We report alignment error rate scores (the lower the better).}
\label{table:word_alignment}
\end{table}

\begin{table*}[t]
\centering
\scalebox{0.9}{
\begin{tabular}{l|cccccccccc|c}
\toprule
Models & En & Cs & De & Fi & Lv & Et & Ro & Hi & Tr & Fr & Avg \\
\midrule
Multilingual NMT &31.64&	27.61&	40.72&	31.88&	31.61&	25.92&	24.25&	32.82&	31.72&	35.58&	31.38
 \\
\xlmt{} &32.71&	33.34&	39.51&	35.52&	33.27&	25.09&	21.79&	37.82&	32.86&	36.21& \bf 32.81
\\
\bottomrule
\end{tabular}}
\caption{Analysis of unsupervised dependency parsing performance on Universal Dependencies. The evaluation metric is UAS F1 score (\%).}
\label{table:dependency}
\end{table*}

\paragraph{Results}

Table~\ref{table:word_alignment} summarizes the performance of multilingual NMT and \xlmt{}. The scores are lower-the-better.
We also report the score of XLM-R for the reference. Both multilingual NMT and \xlmt{} outperform XLM-R because MT data benefits the word alignment.
Compared \xlmt{} with the baseline, it shows that there are significant gains in En-Cs, En-Fr, and En-Ro, indicating much higher similarities of \xlmt{} between two translated words in these languages.
In general, the average alignment error rate across different languages for \xlmt{} achieves 24.56\%, outperforming the multilingual baseline by 1.84\%. This supports our assumption that \xlmt{} improves the similarities of the encoder representations between two languages.

\begin{table}[t]
\centering
\small
\scalebox{0.9}{
\begin{tabular}{l|ccccc|c}
\toprule
Models & En & De & Hi & Tr & Fr & Avg \\
\midrule
XLM-R &85.8 &79.3 &72.8 &76.2 &79.4 &78.7 \\
\midrule
Multilingual NMT &77.1 &73.4 &66.6 &69.7 &72.6 &71.9 \\
\xlmt{} &80.4 &75.2 &66.7 &74.0 &75.3& \bf 74.3\\
\bottomrule
\end{tabular}}
\caption{Analysis of multilingual classification on XNLI. The evaluation metric is accuracy (\%).}
\label{table:xnli}
\end{table}

\subsection{Unsupervised Dependency Parsing}

Prior work~\cite{transformerencoder} prove that the encoder of Transformer-based NMT learns some syntactic information. We investigate that whether \xlmt{} can induce better syntactic tree structures. The self-attention insides Transformer computes the weights between pairs of tokens, which can be formulated as a weighted graph. Therefore, we extract a tree structure from the graph. We compare the extracted tree with its annotated dependency tree to see whether \xlmt{} improves the ability of unsupervised dependency parsing.

\paragraph{Setup}

We compare the accuracy of dependency parsing between multilingual NMT baseline and \xlmt{}. Both models are trained with the WMT-10 dataset in the many-to-many setting. We use Universal Dependencies\footnote{https://universaldependencies.org} as the test set to probe the performance and evaluate in 10 languages (i.e., English, French, Czech, German, Finnish, Latvian, Estonian, Romanian, Hindi, and Turkish) that appear in both WMT-10 and Universal Dependencies. To extract dependency trees, we average the attention scores overall heads in each layer as the weights and compute the maximum spanning trees with Chu-Liu/Edmonds' Algorithm. Since the sentence is tokenized with SentencePiece, we average the weights of all tokens for each word. The gold root of each sentence is used as the starting node for the maximum spanning tree algorithm. We compute the Unlabeled Attachment Score (UAS) with CoNLL 2017's evaluation script\footnote{http://universaldependencies.org/conll17/evaluation.html}.

\paragraph{Results}

As shown in Table~\ref{table:dependency}, we compare the UAS F1 score of multilingual NMT and \xlmt{}. 
We evaluate the performance of each layer and summarize the results of the layer with the highest average score over all languages.
According to Table~\ref{table:dependency}, 
of all 10 languages, the multilingual baseline outperforms \xlmt{} in 3 languages (De, Et, Ro), while \xlmt{} beats the baseline in the rest 7 languages. For Cs, Fi, and Hi, \xlmt{} has a significant gain of more than 3 points compared with the baseline. Generally, \xlmt{} gets 32.81\% UAS, improving the baseline by 1.43\%. This proves that \xlmt{} induces a better syntactic tree structure across different languages, which potentially improves multilingual NMT.

\subsection{Multilingual Classification}

Since multilingual NMT uses a shared representation for different languages, we assume \xlmt{} benefits multilingual NMT by improving the multilingual representations. To verify this, we use the XNLI dataset, which is a widely used testbed for multilingual representation. We evaluate the performance of each language separately. 

\paragraph{Setup}

We compare the accuracy of XNLI between multilingual NMT baseline and \xlmt{}. Both models are trained with the WMT-10 dataset. We retain the encoders and put a projection layer on the top of the first token. The premise and hypothesis are concatenated as the input and fed into the model to produce a label indicating whether there is an entailment, contradiction, or neutral relationship. We fine-tune with the training data of each language. We evaluate the performance in 5 languages (i.e., English, German, Hindi, Turkish, French) that are shared by WMT-10 and XNLI.

\paragraph{Results}

Table~\ref{table:xnli} reports the results on the XNLI dataset. 
XLM-R is the best, showing fine-tuning with MT data degrades the performance on XNLI. This is because the training objective biases towards translation. 
It shows that \xlmt{} beats the multilingual baseline in 4 languages with significant gains (En +3.3\%, De +1.8\%, Tr +4.3\%, Fr +2.7\%) as well as slightly better accuracy in Hi (+0.1\%). The average accuracy across 5 languages is 74.2\%, improving the baseline by 2.3\%.
The results indicate that \xlmt{} improves the representations among different languages, which is important for multilingual NMT, especially when translating low-resource languages.

\section{Related Work}

\paragraph{Multilingual Machine Translation}

\citet{multiway} proposed a many-to-many model to support translating between multiple languages by using specific encoders and decoders for each language while sharing the attention mechanism. \citet{haea2016} and \citet{googlemnmt} introduced a unified model that shared the encoders, decoders, and the attention mechanism for all languages. They used a language token to indicate which target language to be translated. \citet{zeroresource} proved that this multilingual NMT model can generalize to untrained language pairs, which enabled zero-resource translation. \citet{lowresource} showed that training on high-resource languages helps transfer to low-resource machine translation.

More recent work focused on model architecture with different strategies of sharing parameters or representations.
\citet{blackwood-etal-2018-multilingual} proposed to share all parameters but that of the attention layers.
\citet{PlataniosSNM18} introduced a model that learns to generate specific parameters for a language pair while sharing the rest parameters.
\citet{GuHDL18} utilized a transfer-learning approach to share lexical and sentence level representations across multiple source languages into one target language.
In contrast, we do not modify the architecture of multilingual machine translation.

Recently, there are some work focusing on scaling up multilingual machine translation. \citet{mmnmt} performed extensive experiments in
training massively multilingual NMT models, enabling the translation of up to 102 languages within a single model. \citet{opus100} set up a benchmark collected from OPUS for massively multilingual machine translation research and experiments.
Gpipe~\cite{gpipe} scaled up multilingual NMT with a very large and deep Transformer model.
Gshard~\cite{gshard} enabled to scale up multilingual NMT model with Sparsely-Gated Mixture-of-Experts beyond 600 billion parameters using automatic sharding. M2M-100~\cite{m2m} built a multilingual parallel dataset through large-scale mining. They also investigated the methods to increase model capacity through a combination of dense scaling and language-specific sparse
parameters. Different from these work, we do not scale the training data or increase the model size. Instead, we propose to leverage a pretrained model that has been learned on large-scale monolingual data.

\paragraph{Language Model Pre-training}

\citet{bert} and \citet{roberta} use masked language modeling to pretrain the model on large-scale monolingual corpora and transferred to various downstream datasets.
\citet{xlnet} proposed a generalized auto-aggressive pre-training method that enables learning bidirectional contexts by maximizing the expected likelihood over all permutations of the factorization order.
UniLM~\cite{unilm,unilmv2} are unified pretrained language models that can be fine-tuned for both natural language understanding and generation tasks.
\citet{effectiveness:bert} show that language model pre-training provides a good initial point for NLP tasks, which improves performance and generalization capability .
In addition, XLM~\cite{xlm}, XLM-R~\cite{xlmr} and InfoXLM~\cite{infoxlm} are the multilingual pretrained language models that achieve significant gains for a wide range of cross-lingual tasks.
There are some models~\cite{mass,t5,mt5,bart,mbart, multilingualfinetune} based on the encoder-decoder framework that enables fine-tuning the whole models for language generation tasks.
\citet{mrasp} pretrain the multilingual machine translation models with a code-switching objective function.
Compared with previous work, we focus on how to fine-tune pretrained cross-lingual encoders towards multilingual machine translation.

\section{Conclusion}

In this work, we propose \xlmt{} to scale up multilingual machine translation using pretrained cross-lingual encoders. 
This is achieved by initializing the multilingual NMT model with the off-the-shelf XLM-R model. 
\xlmt{} can achieve significant improvements on two large-scale multilingual translation benchmarks, even over the strong baseline with back-translation.
We perform three probing tasks for \xlmt{}, including word alignment, unsupervised dependency parsing, and multilingual classification. 
The probing results explain its effectiveness for machine translation.
This simple method can be used as a new strong baseline for future multilingual NMT systems.

\bibliography{anthology,acl2020}
\bibliographystyle{acl_natbib}

\input{appendix}

\end{document}

%% file: appendix.tex
\appendix

\section{Dataset Statistics}

Table~\ref{table:wmt10} lists the statistics of 10 language pairs from WMT-10. The monolingual data is back-translated as the augmented training data. WMT provides various resources of training data for each language pair. We use all data except Wikititles following~\cite{zcode}.

Table~\ref{table:opus-stats} summarizes the number of training, validation, and test samples for each language from OPUS-100. We remove 5 languages without any validation or test example.

\section{Results on OPUS-100}

We provide the test BLEU of the multilingual baseline and \xlmt{} for all 94 language pairs on OPUS-100 test sets.
Table~\ref{table:opus-x2e} reports the scores on X $\rightarrow$ En test sets. Table~\ref{table:opus-e2x} is on En $\rightarrow$ X test sets.

\begin{table*}[t]
\centering
\small
\begin{tabular}{ccccccc}
\toprule
Code & Language & \#Bitext & \#Mono & Training & Valid & Test \\
\midrule
Fr & French & 10M & 5.0M & WMT15 & Newstest13 & Newstest15 \\
Cs & Czech & 10M & 5.0M & WMT19 & Newstest16 & Newstest18 \\
De & German & 4.6M & 5.0M & WMT19 & Newstest16 & Newstest18 \\
Fi & Finnish & 4.8M & 5.0M & WMT19 & Newstest16 & Newstest18 \\
Lv & Latvian & 1.4M & 5.0M & WMT17 & Newsdev17 & Newstest17 \\
Et & Estonian & 0.7M & 5.0M & WMT18 & Newsdev18 & Newstest18 \\
Ro & Romanian & 0.5M & 5.0M & WMT16 & Newsdev16 & Newstest16 \\
Hi & Hindi & 0.26M & 5.0M & WMT14 & Newsdev14 & Newstest14 \\
Tr & Turkish & 0.18M & 5.0M & WMT18 & Newstest16 & Newstest18 \\
Gu & Gujarati & 0.08M & 5.0M & WMT19 & Newsdev19 & Newstest19 \\
\bottomrule
\end{tabular}
\caption{Statistics and sources of the training, validation, and test sets from WMT. The languages are ranked with the size of parallel corpus.}
\label{table:wmt10}
\end{table*}

\begin{table*}[t]
\centering
\small
\begin{tabular}{llrrrp{1cm}llrrr}
\toprule
Code & Language & Train & Valid & Test &  & Code & Language & Train & Valid & Test \\
af & Afrikaans & 275512 & 2000 & 2000 & & lv & Latvian & 1000000 & 2000 & 2000 \\
am & Amharic & 89027 & 2000 & 2000 & & mg & Malagasy & 590771 & 2000 & 2000 \\
ar & Arabic & 1000000 & 2000 & 2000 & & mk & Macedonian & 1000000 & 2000 & 2000 \\
as & Assamese & 138479 & 2000 & 2000 & & ml & Malayalam & 822746 & 2000 & 2000 \\
az & Azerbaijani & 262089 & 2000 & 2000 & & mr & Marathi & 27007 & 2000 & 2000 \\
be & Belarusian & 67312 & 2000 & 2000 & & ms & Malay & 1000000 & 2000 & 2000 \\
bg & Bulgarian & 1000000 & 2000 & 2000 & & mt & Maltese & 1000000 & 2000 & 2000 \\
bn & Bengali & 1000000 & 2000 & 2000 & & my & Burmese & 24594 & 2000 & 2000 \\
br & Breton & 153447 & 2000 & 2000 & & nb & Norwegian Bokmål & 142906 & 2000 & 2000 \\
bs & Bosnian & 1000000 & 2000 & 2000 & & ne & Nepali & 406381 & 2000 & 2000 \\
ca & Catalan & 1000000 & 2000 & 2000 & & nl & Dutch & 1000000 & 2000 & 2000 \\
cs & Czech & 1000000 & 2000 & 2000 & & nn & Norwegian Nynorsk & 486055 & 2000 & 2000 \\
cy & Welsh & 289521 & 2000 & 2000 & & no & Norwegian & 1000000 & 2000 & 2000 \\
da & Danish & 1000000 & 2000 & 2000 & & oc & Occitan & 35791 & 2000 & 2000 \\
de & German & 1000000 & 2000 & 2000 & & or & Oriya & 14273 & 1317 & 1318 \\
el & Greek & 1000000 & 2000 & 2000 & & pa & Panjabi & 107296 & 2000 & 2000 \\
eo & Esperanto & 337106 & 2000 & 2000 & & pl & Polish & 1000000 & 2000 & 2000 \\
es & Spanish & 1000000 & 2000 & 2000 & & ps & Pashto & 79127 & 2000 & 2000 \\
et & Estonian & 1000000 & 2000 & 2000 & & pt & Portuguese & 1000000 & 2000 & 2000 \\
eu & Basque & 1000000 & 2000 & 2000 & & ro & Romanian & 1000000 & 2000 & 2000 \\
fa & Persian & 1000000 & 2000 & 2000 & & ru & Russian & 1000000 & 2000 & 2000 \\
fi & Finnish & 1000000 & 2000 & 2000 & & rw & Kinyarwanda & 173823 & 2000 & 2000 \\
fr & French & 1000000 & 2000 & 2000 & & se & Northern Sami & 35907 & 2000 & 2000 \\
fy & Western Frisian & 54342 & 2000 & 2000 & & sh & Serbo-Croatian & 267211 & 2000 & 2000 \\
ga & Irish & 289524 & 2000 & 2000 & & si & Sinhala & 979109 & 2000 & 2000 \\
gd & Gaelic & 16316 & 1605 & 1606 & & sk & Slovak & 1000000 & 2000 & 2000 \\
gl & Galician & 515344 & 2000 & 2000 & & sl & Slovenian & 1000000 & 2000 & 2000 \\
gu & Gujarati & 318306 & 2000 & 2000 & & sq & Albanian & 1000000 & 2000 & 2000 \\
ha & Hausa & 97983 & 2000 & 2000 & & sr & Serbian & 1000000 & 2000 & 2000 \\
he & Hebrew & 1000000 & 2000 & 2000 & & sv & Swedish & 1000000 & 2000 & 2000 \\
hi & Hindi & 534319 & 2000 & 2000 & & ta & Tamil & 227014 & 2000 & 2000 \\
hr & Croatian & 1000000 & 2000 & 2000 & & te & Telugu & 64352 & 2000 & 2000 \\
hu & Hungarian & 1000000 & 2000 & 2000 & & tg & Tajik & 193882 & 2000 & 2000 \\
id & Indonesian & 1000000 & 2000 & 2000 & & th & Thai & 1000000 & 2000 & 2000 \\
ig & Igbo & 18415 & 1843 & 1843 & & tk & Turkmen & 13110 & 1852 & 1852 \\
is & Icelandic & 1000000 & 2000 & 2000 & & tr & Turkish & 1000000 & 2000 & 2000 \\
it & Italian & 1000000 & 2000 & 2000 & & tt & Tatar & 100843 & 2000 & 2000 \\
ja & Japanese & 1000000 & 2000 & 2000 & & ug & Uighur & 72170 & 2000 & 2000 \\
ka & Georgian & 377306 & 2000 & 2000 & & uk & Ukrainian & 1000000 & 2000 & 2000 \\
kk & Kazakh & 79927 & 2000 & 2000 & & ur & Urdu & 753913 & 2000 & 2000 \\
km & Central Khmer & 111483 & 2000 & 2000 & & uz & Uzbek & 173157 & 2000 & 2000 \\
kn & Kannada & 14537 & 917 & 918 & & vi & Vietnamese & 1000000 & 2000 & 2000 \\
ko & Korean & 1000000 & 2000 & 2000 & & wa & Walloon & 104496 & 2000 & 2000 \\
ku & Kurdish & 144844 & 2000 & 2000 & & xh & Xhosa  & 439671 & 2000 & 2000 \\
ky & Kyrgyz & 27215 & 2000 & 2000 & & yi & Yiddish & 15010 & 2000 & 2000 \\
li & Limburgan & 25535 & 2000 & 2000 & & zh & Chinese & 1000000 & 2000 & 2000 \\
lt & Lithuanian & 1000000 & 2000 & 2000 & & zu & Zulu & 38616 & 2000 & 2000 \\
\bottomrule
\end{tabular}
\caption{Statistics of the training, validation, and test sets from OPUS-100. The languages are ranked in alphabet order.}
\label{table:opus-stats}
\end{table*}

\begin{table*}[t]
\centering
\begin{tabular}{l|cccccccccc}
\midrule
Code & af & am & ar & as & az & be & bg & bn & br & bs \\
\midrule
Multilingual NMT & 51.8 & 23.0 & 36.0 & 55.7 & 27.2 & 28.4 & 32.1 & 21.9 & 23.3 & 30.8 \\
\xlmt{} & 53.2 & 22.5 & 37.8 & 58.3 & 26.6 & 28.7 & 32.5 & 23.2 & 23.5 & 31.0 \\
\midrule
Code & ca & cs & cy & da & de & el & eo & es & et & eu \\
\midrule
Multilingual NMT & 38.0 & 34.1 & 48.8 & 36.3 & 33.8 & 32.3 & 37.9 & 39.5 & 35.8 & 20.1 \\
\xlmt{} & 38.7 & 34.8 & 49.8 & 37.1 & 34.9 & 33.5 & 38.3 & 40.9 & 36.2 & 20.6 \\
\midrule
Code & fa & fi & fr & fy & ga & gd & gl & gu & ha & he \\
\midrule
Multilingual NMT & 22.9 & 24.5 & 33.9 & 42.5 & 61.5 & 75.4 & 30.6 & 59.8 & 24.1 & 34.4 \\
\xlmt{} & 23.6 & 25.3 & 34.6 & 40.6 & 63.3 & 77.7 & 31.0 & 61.6 & 24.1 & 36.0 \\
\midrule
Code & hi & hr & hu & id & ig & is & it & ja & ka & kk \\
\midrule
Multilingual NMT & 27.5 & 31.0 & 26.7 & 33.8 & 53.9 & 23.4 & 35.7 & 14.0 & 22.4 & 28.7 \\
\xlmt{} & 28.4 & 31.9 & 28.6 & 34.6 & 55.1 & 24.2 & 36.1 & 14.8 & 22.9 & 29.1 \\
\midrule
Code & km & kn & ko & ku & ky & li & lt & lv & mg & mk \\
\midrule
Multilingual NMT & 37.5 & 41.2 & 15.0 & 24.8 & 39.0 & 36.4 & 41.9 & 45.5 & 28.1 & 34.0 \\
\xlmt{} & 37.2 & 43.6 & 15.6 & 26.0 & 41.6 & 37.9 & 43.7 & 46.3 & 29.0 & 35.0 \\
\midrule
Code & ml & mr & ms & mt & my & nb & ne & nl & nn & no \\
\midrule
Multilingual NMT & 18.9 & 50.7 & 29.7 & 62.3 & 19.5 & 43.3 & 46.9 & 31.3 & 37.0 & 25.0 \\
\xlmt{} & 19.2 & 52.3 & 30.0 & 63.0 & 20.7 & 44.6 & 47.4 & 32.1 & 37.8 & 25.6 \\
\midrule
Code & oc & or & pa & pl & ps & pt & ro & ru & rw & se \\
\midrule
Multilingual NMT & 16.4 & 33.5 & 45.5 & 26.4 & 38.5 & 36.8 & 37.5 & 33.8 & 28.3 & 16.0 \\
\xlmt{} & 15.3 & 35.1 & 46.2 & 27.7 & 41.6 & 37.3 & 39.0 & 35.1 & 28.4 & 14.8 \\
\midrule
Code & sh & si & sk & sl & sq & sr & sv & ta & te & tg \\
\midrule
Multilingual NMT & 55.2 & 22.2 & 38.2 & 27.7 & 41.7 & 30.3 & 31.9 & 29.1 & 43.2 & 24.6 \\
\xlmt{} & 56.4 & 23.5 & 39.1 & 28.4 & 43.1 & 31.5 & 32.9 & 30.1 & 43.8 & 24.3 \\
\midrule
Code & th & tk & tr & tt & ug & uk & ur & uz & vi & wa \\
\midrule
Multilingual NMT & 21.1 & 48.5 & 24.6 & 19.9 & 20.8 & 27.0 & 21.5 & 20.2 & 25.2 & 31.2 \\
\xlmt{} & 21.9 & 49.1 & 25.1 & 20.3 & 20.7 & 28.0 & 21.6 & 18.7 & 26.2 & 33.3 \\
\midrule
Code & xh & yi & zh & zu &  &  &  &  &  &  \\
\midrule
Multilingual NMT & 24.6 & 27.3 & 37.8 & 50.4 &  &  &  &  &  &  \\
\xlmt{} & 26.5 & 29.9 & 39.0 & 50.5 &  &  &  &  &  &  \\

\midrule
\end{tabular}
\caption{X $\rightarrow$ En test BLEU for 94 language pairs in many-to-many setting on the OPUS-100 test sets. The languages are ranked in alphabet order.}
\label{table:opus-x2e}
\end{table*}

\begin{table*}[t]
\centering
\begin{tabular}{l|cccccccccc}
\midrule
Code & af & am & ar & as & az & be & bg & bn & br & bs \\
\midrule
Multilingual NMT & 45.1 & 18.7 & 20.0 & 41.5 & 28.5 & 26.2 & 28.8 & 11.6 & 25.0 & 21.5 \\
\xlmt{} & 44.6 & 21.2 & 20.4 & 41.9 & 27.9 & 26.5 & 29.8 & 11.4 & 25.2 & 21.9 \\
\midrule
Code & ca & cs & cy & da & de & el & eo & es & et & eu \\
\midrule
Multilingual NMT & 35.2 & 26.7 & 42.2 & 34.8 & 30.1 & 26.7 & 33.6 & 37.0 & 30.2 & 14.1 \\
\xlmt{} & 35.8 & 26.6 & 44.1 & 35.4 & 30.7 & 27.3 & 34.3 & 37.4 & 29.8 & 14.3 \\
\midrule
Code & fa & fi & fr & fy & ga & gd & gl & gu & ha & he \\
\midrule
Multilingual NMT & 9.6 & 20.9 & 32.4 & 33.1 & 50.4 & 27.6 & 27.3 & 52.4 & 47.5 & 28.2 \\
\xlmt{} & 9.4 & 21.2 & 32.9 & 34.5 & 51.1 & 31.6 & 27.8 & 52.5 & 48.6 & 28.8 \\
\midrule
Code & hi & hr & hu & id & ig & is & it & ja & ka & kk \\
\midrule
Multilingual NMT & 19.8 & 24.2 & 20.3 & 30.0 & 45.8 & 21.1 & 29.4 & 12.0 & 16.6 & 25.1 \\
\xlmt{} & 20.9 & 24.7 & 20.3 & 30.3 & 45.7 & 20.8 & 30.5 & 12.3 & 17.4 & 25.0 \\
\midrule
Code & km & kn & ko & ku & ky & li & lt & lv & mg & mk \\
\midrule
Multilingual NMT & 19.6 & 28.6 & 6.0 & 8.0 & 33.4 & 32.3 & 35.4 & 39.5 & 22.4 & 33.3 \\
\xlmt{} & 20.2 & 29.4 & 6.7 & 7.9 & 35.1 & 31.5 & 36.2 & 40.1 & 22.6 & 33.9 \\
\midrule
Code & ml & mr & ms & mt & my & nb & ne & nl & nn & no \\
\midrule
Multilingual NMT & 5.1 & 31.8 & 24.2 & 47.4 & 13.0 & 37.7 & 42.4 & 27.2 & 30.7 & 28.5 \\
\xlmt{} & 5.7 & 33.5 & 24.5 & 48.0 & 11.4 & 38.6 & 42.3 & 27.9 & 30.6 & 28.9 \\
\midrule
Code & oc & or & pa & pl & ps & pt & ro & ru & rw & se \\
\midrule
Multilingual NMT & 24.2 & 34.1 & 43.6 & 20.8 & 41.6 & 31.6 & 31.0 & 28.4 & 69.4 & 25.7 \\
\xlmt{} & 23.3 & 31.5 & 42.9 & 21.3 & 41.9 & 32.3 & 31.3 & 28.7 & 68.8 & 26.2 \\
\midrule
Code & sh & si & sk & sl & sq & sr & sv & ta & te & tg \\
\midrule
Multilingual NMT & 50.9 & 10.6 & 29.7 & 24.2 & 37.0 & 20.8 & 31.0 & 18.8 & 32.2 & 28.8 \\
\xlmt{} & 51.4 & 10.5 & 30.3 & 25.2 & 37.4 & 21.5 & 31.7 & 19.4 & 32.3 & 28.9 \\
\midrule
Code & th & tk & tr & tt & ug & uk & ur & uz & vi & wa \\
\midrule
Multilingual NMT & 8.7 & 45.4 & 16.7 & 19.6 & 12.0 & 15.6 & 19.6 & 15.2 & 21.9 & 27.5 \\
\xlmt{} & 9.1 & 45.1 & 17.1 & 20.2 & 12.4 & 16.7 & 19.7 & 16.3 & 22.1 & 29.4 \\
\midrule
Code & xh & yi & zh & zu &  &  &  &  &  &  \\
\midrule
Multilingual NMT & 14.0 & 27.9 & 41.1 & 35.5 &  &  &  &  &  &  \\
\xlmt{} & 13.6 & 27.2 & 41.5 & 36.3 &  &  &  &  &  &  \\
\midrule

\end{tabular}
\caption{En $\rightarrow$ X test BLEU for 94 language pairs in many-to-many setting on the OPUS-100 test sets. The languages are ranked in alphabet order.}
\label{table:opus-e2x}
\end{table*}